\documentclass[twocolumn,pre]{revtex4}

\usepackage{graphicx}
\usepackage{epstopdf}
\usepackage{color}
\usepackage{footnote}
\usepackage{amsmath}
\usepackage{natbib}
\usepackage{multirow}
\usepackage{graphicx}
\usepackage{ulem}

\newcommand{\lesmis}{Les Mis\'erables}
\newcommand{\setT}{\mathcal{T}}
\newcommand{\setD}{\mathcal{D}}
\newcommand{\setW}{\mathcal{W}}

\newcommand{\dimT}{|\setT|}
\newcommand{\dimD}{|\setD|}
\newcommand{\dimW}{|\setW|}

\newcommand{\Chs}{\mathcal{C}}
\newcommand{\psen}{\pi}
\newcommand{\nsen}{\nu}
\newcommand{\sen}{\sigma}

\renewcommand{\emph}{\textit}

\newcommand{\etc}{\textit{etc.}}

\newcommand{\set}[1]{\{#1\}} 
\newcommand{\kw}[1]{\texttt{#1}}

\newcommand{\defn}[1]{\textbf{#1}}

\newcommand{\logten}{\log_{10}}

\newlength{\figurewidth}
\ifdim\columnwidth<20cm
\else
\fi
\begin{document}

\title{Mapping Out Narrative Structures and Dynamics Using Networks and Textual Information}
\author{Semi Min}
\affiliation{Graduate School of Culture Technology, KAIST, Daejeon, Korea 34141}
\author{Juyong Park}
\affiliation{Graduate School of Culture Technology, KAIST, Daejeon, Korea 34141}

\begin{abstract}
Human communication is often executed in the form of a narrative, an account of connected events composed of characters, actions, and settings. A coherent narrative structure is therefore a requisite for a well-formulated narrative -- be it fictional or nonfictional -- for informative and effective communication, opening up the possibility of a deeper understanding of a narrative by studying its structural properties. In this paper we present a network-based framework for modeling and analyzing the structure of a narrative, which is further expanded by incorporating methods from computational linguistics to utilize the narrative text. Modeling a narrative as a dynamically unfolding system, we characterize its progression via the growth patterns of the character network, and use sentiment analysis and topic modeling to represent the actual content of the narrative in the form of interaction maps between characters with associated sentiment values and keywords.  This is a network framework advanced beyond the simple occurrence-based one most often used until now, allowing one to utilize the unique characteristics of a given narrative to a high degree.  Given the ubiquity and importance of narratives, such advanced network-based representation and analysis framework may lead to a more systematic modeling and understanding of narratives for social interactions, expression of human sentiments, and communication.
\end{abstract}
\maketitle

\section{Introduction}
\label{sec:intro}
Recent advances in quantitative methodologies for the modeling and analyses of large-scale heterogeneous data have enabled novel understanding of various complex systems from the social, technological, and biological domains~\cite{michel2011quantitative}.  The field of application is also rapidly expanding, now including the traditional academic fields of cultural studies humanities. It is allowing researchers to obtain novel answers to both long-standing problems by finding complex patterns that were previously hidden. Recent examples include high-throughput analyses of language and literature based on massive digitization of books (e.g., Project Gutenberg~\cite{gutenberg} and Google Books) and proliferation of social media~\cite{michel2011quantitative,dodds2015human}, emergent processes in cultural history~\cite{schich2014network}, and scientific analysis of art~\cite{kim2014large}. 

A theoretical data modeling and analysis framework that has attracted attention for cultural studies is networks~\cite{schich2014network,park2015topology}. Network science attempts to understand the structure and behavior of a complex system from the connection and interaction patterns between its components~\cite{Newman:2010fk,AlBara01,easley2010networks,han2011data}.  Owing to its flexibility as a modeling framework, network science has led to a novel understanding of many systems that are not only easily recognizable as a network such as the Worldwide Web~\cite{adamic2000power, albert1999internet}, the Internet~\cite{choi2006comparing}, but also those that have been extensively studied in non-network contexts such as biological systems or social organizations~\cite{borgatti2003network,grimm2005pattern}.

In this paper we propose a network science-based framework for a cultural system that is ubiquitous in society and boasts a long history of study but we believe still can benefit from one: Narratives. Narratives (or stories) are important in that they are the most common way in which we communicate and recount our experiences.  The connection between networks and narratives can also be seen in the very definition of the word: The New Oxford American Dictionary, for instance, defines narrative as ``a spoken or written account of connected events.''  This suggests that using networks may help us understanding how the various building blocks of narratives are weaved to become a coherent structure for effective delivery of messages and arousal of emotions.  This way of thinking about narratives is deeply correlated with an interesting recent movement in literary studies named ``distant reading'' proposed by Moretti~\cite{moretti2011network,moretti2013distant,morettipamphlets}. Distant reading is an approach to literature based on processing large amounts of literary data to devise and construct general ``models'' of narratives to understand them as a class, in contrast to reading each work very closely (hence the term ``distant'') to understand it.  A model constructed through reduction and abstraction, the reasoning goes, would enable us to grasp the general underlying structures and patterns of a class of complex objects called narrative, as an X-ray machine would allow us to understand the general skeletal features of the human body.

To many of us this way of thinking is familiar as the very principle of research in the natural sciences: To understand a system, one collects data and performs statistical analysis based on abstract models to gain an understanding of the general characteristics of the system.  A model of a system has the following characteristics.  An abstract representation or notion of a system, a model necessarily excludes some features of the system it is representing.  A random exclusion of features, of course, is unlikely to result in a useful model; it is important to make a judicious choice on which features to retain and which to exclude so that the model incorporates important or essential of the system. Of course, it is very difficult to know beforehand which is the best choice of features. One practical starting point can be a common description of the system by people, since such a description is already a type of mental representation which can be viewed as a model, however rudimentary.  The network model of a narrative, from this perspective, appears to be sensible and immediately understandable; in many instances when we recount a story, we focus prominently on the characters and their relationships. Take the Star Wars movie franchise, for instance, the top grossing space opera in modern times~\cite{boxofficemojo,thenumbers}.   Once the generic physical setting of ``a galaxy far, far away'' is presented, the story progresses via the character's actions, adventures, and relationships and interactions with others; that Leia Organa and Luke Skywalker are twins play an crucial role in their fate, and the revelation via ``I am your father'' is perhaps the most memorable scene in the narrative.  In addition to these individual relationships, group-level relationships are important as well for the story, such as the Empire versus the Rebel Alliance, the dark side versus the light side of the Force,~\etc~Examples abound from history: The story of Oedipus that precedes Star Wars in terms of shocking familial revelation; \textit{Dexter}, a favorite American TV show of one of the authors, is a series of episodes that portray the titular character navigating his social world of his sibling, family, and rival criminals~\footnote{As a mature-rated show, the titular character's identity and the overarching plot of the drama may be discomforting and too cruel for some readers to describe here; we refer the interested to seek appropriate sources for more information.}

These examples all function as empirical bases for approaching narratives from the network modeling framework.  One of the earliest models proposed for narratives in the distant reading philosophy introduced above, in fact, was networks.  Moretti applied the network framework to Shakespeare's Hamlet for detecting specific regions in the plot, and performed many experiments such as extracting specific nodes in the network of characters to observe changes and make comparisons between different networks.  Other network-based studies of narrative include the study of the community structure of the character network in Victor Hugo's~\lesmis~\cite{Newman:2004}, the social networks of characters based on conversation in 19th-century British novels~\cite{elson2010extracting}, networks of mythologies and sagas~\cite{mac2012universal,mac2013network,kydros2015homer}, and more recently, a technique for dialog detection in novels applied to writer J.~K.~Rowling's Harry Potter series~\cite{waumans2015topology}. While these serve to demonstrate the scientific community's interest in network-based understanding of narratives, most works are limited to the story of the static topological properties of networks found in the said stories, when we know that narratives are essentially dynamically progressing entities, and the text itself is a source of much information that can be used extensively. Given the wide range of analytical and computational tools that constitute network science, we believe there is much opportunity for further studies of narratives using networks that take into account such essential aspects.  This paper is intended to be one such attempt inspired by those works, hopefully laying out potential future directions in utilizing network science and computational linguistics for understanding the dynamics of narratives in a systematic manner.

\section{Materials and Methods}
\label{sec:matemeth}
\subsection{Material: Victor Hugo's~\lesmis}
We analyze Victor Hugo's novel~\lesmis~using the methods introduced in this paper. Set around the popular uprising in Paris in 1832 CE, \lesmis~is known for its vivid depiction of the conditions of the tumultuous times and intuition into the human psyche via multiple intersecting plots involving richly developed characters~\cite{welsh1978opening}. Its main plot follows fugitive Jean Valjean's trajectory that shows him transform into a force for good while being constantly haunted by his criminal past. During his journey he interacts with many characters, some helpful and friendly, and others antagonistic and hostile. The most important characters of the novel include the following:
\begin{itemize}	
	\item \emph{Fantine}: A young woman abandoned with daughter Cosette early in the novel. She later leaves Cosette in the care of the Thenadiers, who then abuse her. She is rescued by Valjean when Javert arrests her on charge of assaulting a man.
	\item \emph{Cosette}: Fantine's daughter, later adopted by Valjean. Under Valjean's care she grows into a beautiful woman, and falls in love with Marius.
	\item \emph{Marius}: A young man associated with the ``Friends of the ABC (\emph{Les Amis de l'ABC} in French),'' a group of revolutionaries.  He is critically wounded at the barricade, but is rescued by Valjean. He later marries Cosette.
	\item \emph{Javert}: A police inspector in a relentless pursuit of Valjean. After being rescued by Valjean at the barricades and realizing the immorality of the old French system he has served loyally, he commits suicide.
	\item \emph{Thenadier}: A wretched man who abuses young Cosette. A lifetime schemer of robbery, fraud, and murder, he conspires to rob Valjean until Marius stops him, and gets arrested by Javert.
\end{itemize}

\subsection{Method: Interacting Timelines and Network Construction}
\label{sec:timeline}
The widely-accepted essential building blocks of a narrative are characters (also called agents or actants), events, and the causal or temporal relationships that weave them together~\cite{rimmon2003narrative, bal2009narratology}.  An interrelated sequence composed of those elements is called a plot which may be viewed as the backbone of a narrative.  A narrative may also be broken down into formal units such as acts, scenes, chapters,~\etc~\cite{abbott2008cambridge}. Historically there have been many attempts to establish a general form of narrative structure, of which a well-known example is Aristotle's three-act plot structure theory. It states that Act One presents the central theme and questions, followed by Acts Two and Three that present major turning points and conclusion. Variant forms exist such as the four-act structure theory~\cite{field2007screenplay,vogler2007writer}.

While these have existed for a long time and been widely applied, we find it difficult to imagine that there is an \emph{a priori} reason for all narratives to consist of three or four parts. Then, can we deduce the structure of a narrative from the narrative itself? It appears that the increasing availability of narrative texts in digital format and analytical methods for data analysis offer an opportunity for a new look at narrative structures, and the formulation of a flexible framework that can properly capture the complexity of a given narrative.

An interesting pair of concepts helpful for picturing the content of a narrative that serve as the basis for formalism was given by Propp~\cite{propp2010morphology} who, while trying to establish a symbolic notation-based formalism for Russian folktales, proposed that narrative content consists of two layers that he labeled the~\defn{fabula} and the~\defn{sjuzet}. The fabula refers to the entire world that contains the narrative, while the sjuzet refers to those elements of that world explicitly presented to the audience. For instance, if the narrative is depicting a man dining with his family in his home, the sjuzet comprises the man and his family (the characters), the act of dining (the event), and his home (the place), while the fabula is all of the above plus the rest of the story world such as the man's colleagues at work, their concurrent actions and whereabouts,~\etc~ The sjuzet therefore can be considered the part of the story world currently under observation, and the rest of the fabula the part that ``operates'' in the background.  Each component of the fabula may or may not become sjuzet (explicitly presented to the audience) at another point in the narrative, but they are nevertheless indispensable for the consistency of the story world and future plot development via implicit action.

\begin{figure*}
\includegraphics[width=18cm]{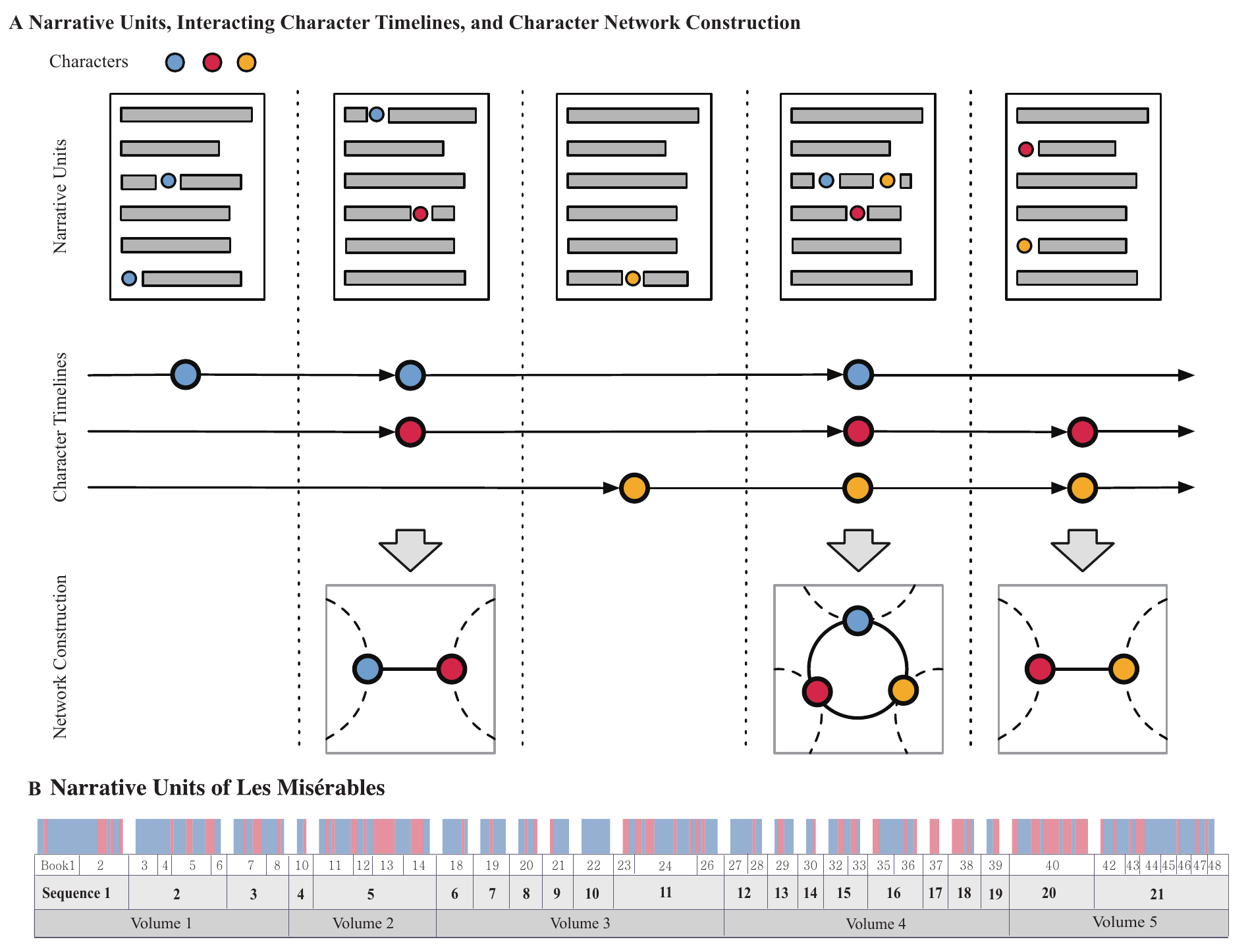} 
\caption{{\bf Interacting Timeline Framework for Network Modeling of Narratives.}
(A) Construction of the character network from a narrative. We represent the narrative as a set of character timelines, the record of appearances of the characters in narrative units (e.g. chapters, scenes,~\etc). An interaction can be defined as co-appearance in a narrative unit.
(B) A narrative unit is not unique. One may use the author's designation (i.e. the Volumes, Books, or Chapters in a novel) or define a new one such as the Sequence based on the unit-to-unit continuity of character compositions (defined in Sec.~\ref{sec:sentimentanalysis}). The narrative units in Victor Hugo's \lesmis~are shown here, from the finest (Chapters, top) to the coarsest (Volumes, bottom).}
\label{fig1}
\end{figure*}

We start by representing a narrative as a set of \defn{character timelines}, basically the record of a character's appearances in the narrative. The point of appearance is marked in narrative units which can be scenes, chapters,~\etc, shown in Fig.~\ref{fig1}. In our paper we follow the convention used in the construction of the character network from Victor Hugo's \lesmis~in Ref.~\cite{Newman:2004}: two characters are connected in the network when they appear in the same narrative unit. An interaction defined in this fashion would be more general than direct conversations, as it could include a common experience or shared space in addition to a conversation.  The narrative format can also present some practical issues in defining an interaction.  In a play or movie script, for instance, it would be much easier to identify a conversation between characters as an interaction, which would be more explicit and narrower in scope. An online resource named \texttt{moviegalaxies} provides a collection of social networks of characters in hundreds of movies built in this way, along with static network properties such as the diameters and clustering coefficients~\cite{moviegalaxies}. Using this narrower definition in a novel is potentially problematic: it is difficult to detect conversations in a novel (though some advances have been recently made~\cite{waumans2015topology}), but more fundamentally it would miss non-verbal interactions which exist abundantly in a novel.  For this reason, it is difficult to state at this point which would be a better approach. Perhaps a comparison study could be illustrating, although it is out of the scope of this work~\cite{morettipamphlets}. The rest of this paper is dedicated to exploring what the character network based on Fig.~\ref{fig1} can tell us about the narrative structure and how it progresses.  The methodology will be demonstrated using the English translation of Victor Hugo's~\lesmis~\cite{hugo1862miserables}, although it will be clear that the formalism itself applicable to any comparable narrative.  Our choice of Hugo's work is based on its stature as a classic known for a set of richly developed characters~\cite{welsh1978opening}, familiarity in network science~\cite{Newman:2004}, and the free availability of the complete text on Project Gutenberg.  Using the original French version would be ideal, but we point out that the network construction according to Fig.~\ref{fig1} is unaffected, and the wider availability of advanced computational linguistic tools for the English language does provide advantages for incorporating the text for enriched analyses that will be demonstrated in the latter part of the work.

In Fig.~\ref{fig1}~(B) we show the narrative units in \lesmis~on several levels. From top to bottom, they are the Chapters (colored according to their Sentiment Polarity Index defined in Sec.~\ref{sec:sentiment}), Books (groups of Chapters), Sequences (groups of Books), and Volume (even larger groups of Books). All but the Sequences, whose definition is given later in Sec.~\ref{sec:sentimentanalysis}, are by the author's designation. It is reasonable to assume that the author intended each unit to represent a theme or subplot.  The five Volumes of~\lesmis, for instance, are titled ``Fantine,'' ``Cosette,'', ``Marius,'' ``The Idyll in the Rue Plumet and the Epic in the Rue St. Denis,'', and ``Jean Valjean,'' indicating their central character or plot.  Since the different narrative units offer a varying degree of resolution of the narrative, one may again choose the one that is most useful for their purposes.  For our goal of studying the complexity of~\lesmis, however, the five Volumes appear too few; we therefore choose to work with the Chapters (of which there are $365$) for most purposes, and the Sequence in a later analysis.

\subsubsection{Network Topology and Growth Patterns}
\label{sec:growth}
From the network of characters built based on the Interacting Timelines of Fig.~\ref{fig1}~(A) we can measure various static network properties.  But a narrative is essentially a dynamical system that unfolds in time; what interests a reader is how the story is told in time, not necessarily the final, static network of characters. We need to study how the network grows over time and what we can learn about the narrative from it. This is because the network growth is essentially coupled to the narrative flow: Starting from an empty network in the beginning of the narrative, the network grows as new characters are introduced and interact with others.  In this sense, we can say that the temporal growth of the network is intimately connected to the concept of the so-called narrative stages.  A common classification of narrative stages includes \emph{Exposition}, \emph{Rising Action}, \emph{Climax}, \emph{Falling Action},  \emph{Resolution},~\etc,~named according to their role and nature~\cite{freytag1896freytag} . For example, the Exposition stage introduces the characters and the space they inhabit. Once the motives and allegiances of the characters are presented, in the Rising Action the characters begin to struggle against each other until all conflicts are resolved through the later stages.

We study the network growth pattern on two levels. First, on the aggregate level, we measure the growth the number of nodes $n$ and edges $m$ of the network. Second, on the individual character level, we measure two values, appearance $a$ (the number of chapters in which a character makes an appearance) and degree $k$ of the characters.

\subsection{Method: Sentiment Analysis and Topic Modeling}
\begin{figure*}
\includegraphics[width=18cm]{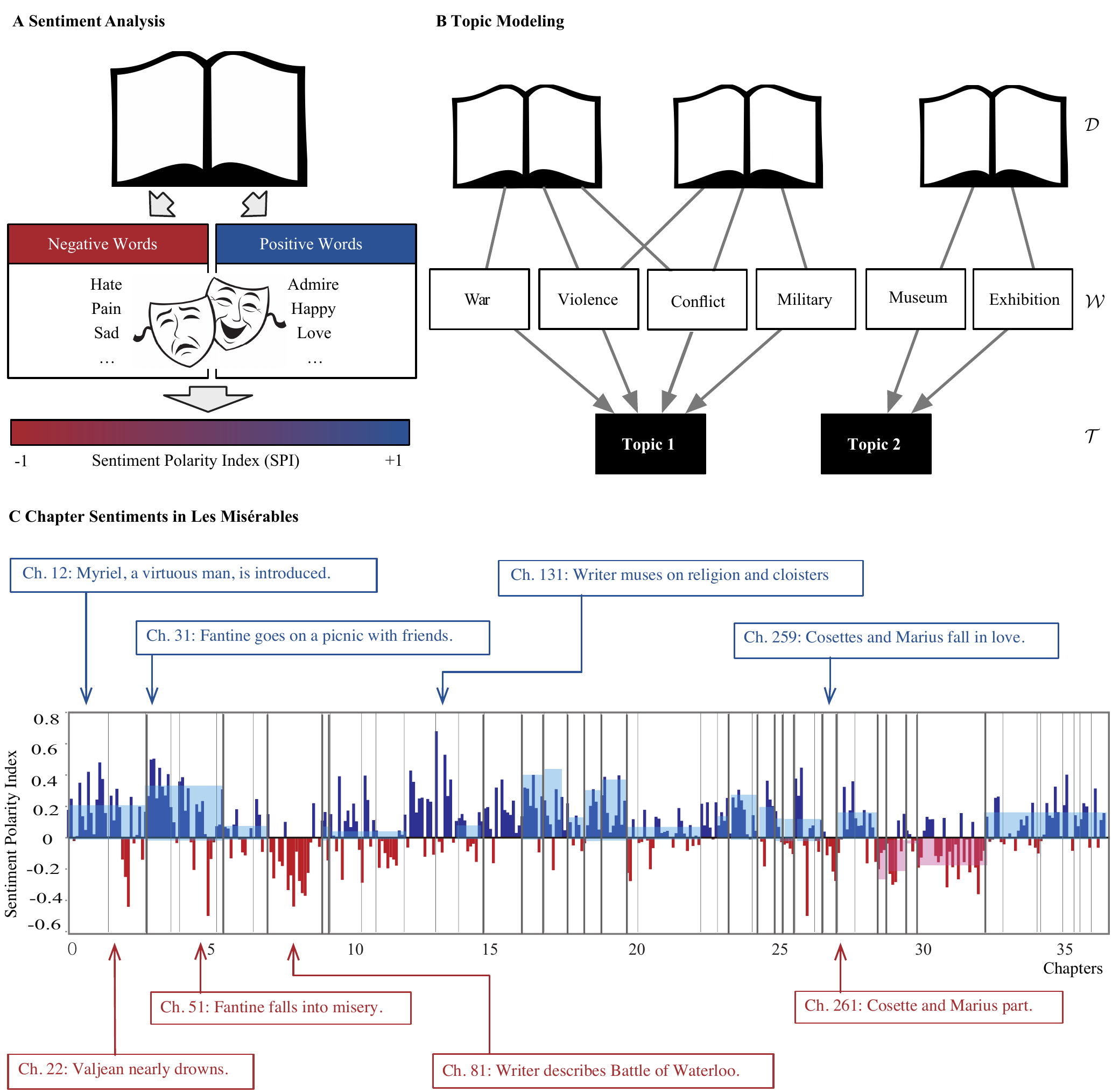}
\caption{{\bf Sentiment Analysis and Topic Modeling of Narratives.}
(A) The principle of sentiment analysis. Words associated with positive or negative sentiments contribute towards the Sentimental Polarity Index (SPI) of the text ranging from $-1$ (most negative) to $+1$ (most positive). (B) The principle of topic modeling.  Clusters of words detected from a set of texts that tend to appear together are identified as the topics. (C) SPIs of the chapters of~\lesmis. Vertical gray bars indicate the 21 Sequences of \lesmis. Each sequence is colored according to the sign of the mean SPI of its constituent chapters (blue for positive, and red for negative). We compare the  SPI and content for eight chapters in the narrative: Positive chapters depict uplifting characters or events (e.g., introduction of Myriel, a man of great character in Chapter 12) and happy events (e.g., Fantine going on a picnic, Cosette and Marius falling in love,~\etc), while negative chapters depict pain and suffering (e.g., Valjean nearly drowning, Fantine in misery, war, lovers parting,~\etc)}
\label{fig2}
\end{figure*}

An analysis focused solely on the network topology leaves out an essential component of a narrative, the text.  This is important because a narrative is in essence much more than a record of who-meets-whom; in the form of text, a narrative contains the details that can vary significantly between interactions~\cite{vogler2007writer,propp2010morphology}.  In \lesmis, for instance, the nature of Valjean's relationships to different characters that is at the center of its drama -- at the same time a savior and protector to Cosette, and a fugitive criminal to Javert -- is wholly missing in the simple appearance-based network.  This means that leveraging the actual text of the narrative may lead to a richer and proper understanding of the narrative, which we perform by using some tools developed in computational linguistics.  Here we utilize two: The first tool is \defn{Sentiment Analysis} that identifies the positive and negative sentimental qualities of a text, which allows us to study the sentimental states of character relationships and the build-up and the resolution of tension in the narrative.  The second tool is \defn{Topic Modeling} that identifies the topics inside the novel, which allow us to associate the characters with the topics at different points in the narrative that define the characters' states, and quantify the impact of events on the characters.

\subsubsection{Sentiment Analysis}
\label{sec:sentiment}
Sentiment Analysis, also called Mood Analysis or Opinion Mining, is a technique for determining the sentimental qualities of a given text based on the words it contains.  Its origin can be traced back to an attempt in the 1990's to translate written reviews of products into numerical rating scores: To this day it is common to produce a numerical \emph{Sentiment Polarity Index} (SPI) of a given text that shows its positive or negative quality.  Basically it count the words of known positive or negative sentimental states from a text to produce SPI.  For instance, words such as ``admire,'' ``happy,'' and ``love'' contribute to the text's positive sentiment, where as ``hate,'' ``pain,'' and ``sad'' would contribute to its negative sentiment. (See Fig.~\ref{fig2}~(A)). We note an interesting connection to the Western literary tradition of the generic division of drama into comedy and tragedy, often stylized using two masks -- the laughing that represents Thalia, the Muse of comedy in Greek and Roman mythology, and weeping one that represent Melpomene the Muse of tragedy, also shown in Fig.~\ref{fig2}~(A). Here we use the LIWC (Linguistic Inquiry and Word Count)~\cite{tausczik2010psychological} program, one of several available~\cite{gonccalves2013comparing}, to determine the SPI of the chapters of \lesmis.  LIWC actually returns two separate values, $\psen\ge0$ and $\nsen\ge0$, for the positive and negative sentiments for the input text, which we combine, for convenience, into a single SPI variable

\begin{eqnarray}
	\sigma\equiv\logten\biggl(\frac{\psen+1}{\nsen+1}\biggr).
\end{eqnarray}

Defined in this way, $\sigma>0$ when the text is net positive $(\psen>\nsen)$, $\sigma=0$ when neutral $(\psen=\nsen)$, and $\sigma<0$ when net negative $(\psen<\nsen)$.  We now have a set of values $\Sigma=\set{\sigma_1,\ldots,\sigma_c}$, where $c=365$ is the number of chapters in~\lesmis.

From $\Sigma=\set{\sigma}$ we can compute SPIs of the characters and character pairs using the timeline framework in Fig.~\ref{fig1}~(A).  If a character $\alpha$, for instance, has appeared in Chapters $1$, $2$, and $100$, we define $\Sigma[\alpha]=\set{\sigma_1,\sigma_2,\sigma_{100}}$ to be the SPI set of $\alpha$, from which we can calculate quantities such as the character's average character SPI, $\overline{\sen}[\alpha] = (\sen_1+\sen_2+\sen_{100})/3$.  The SPI of a character pair is similar: if two characters $\alpha$ and $\beta$ have co-appeared in Chapters $2$ and $100$, for instance, their average SPI is $\overline{\sen}[\alpha,\beta] = (\sen_{2}+\sen_{100})/2$.

\subsubsection{Topic Modeling}
\label{sec:topic}
Our second example of incorporating textual information for network-based narrative study is topic modeling.  We will see that it allows us to determine the ``topical state'' of a character at any given point in the narrative and map out a detailed picture of interaction between characters.  Topic modeling is a method for extracting clusters of correlated keywords from a set of documents that can be identified as separate ``topics'' of the texts.  The basic idea is presented in Fig.~\ref{fig2}~(B) through a tripartite network composed of three layers of nodes: $\setD$ of documents, $\setW$ of words, and $\setT$ of topics.  The goal is to find $\setT$, essentially ``bags of words'' appearing often together in documents, from the text data consisting of $\setD$ and $\setW$.  Many studies have reported the success of topic modeling in identifying word sets that match the human understanding of groups of texts, and its practical applicability to problems like word sense induction~\cite{jurgens2010s,van2011latent,stevens2012exploring}. 

Here we employ the Non-Negative Matrix Factorization (NNMF), famously used for identifying distinguishable parts in images~\cite{lee1999learning,lee2001algorithms,xu2003document,zhao2004empirical}. It decomposes the word--document TF-IDF (Term-Frequency -- Inverse Document Frequency) matrix $M$ ($\dim(M)=\dimW\times\dimD$) into $QH$, product of two matrices $Q$ and $H$ such that $\dim(Q)=\dimW\times\dimT$ and $\dim(Q)=\dimT\times\dimD$.  The number of topics $\dimT$ is an input parameter typically set to be smaller than $\dimW$ and $\dimD$~\footnote{The decomposition is also approximate in practice, i.e. $M\simeq QH$ with the difference between $M$ and $QH$ (called reconstruction error) measured by the squared Frobenius norm.}. We can then interpret the matrix $Q=\set{q_{ij}}$ as the association strength between word $i$ and topic $j$, and $H=\set{h_{jk}}$ as that between topic $j$ and chapter $k$.  Using the framework of Fig.~\ref{fig1}~(A) we can again define the character-topic association strength $t_{\alpha k}$ between character $\alpha$ and topic $k$ as follows:

\begin{eqnarray}
	t_{\alpha k} \equiv \frac{\sum_{\Chs_{\alpha}} h_{jk}}{\sum_{k\in\setT,j\in\Chs_{\alpha}} h_{jk}},
\end{eqnarray}
where $\Chs_{\alpha}$ is the set of chapters where character $\alpha$ appears in. It is the normalized sum of all topic-chapters associations $h_{jk}$ from the chapters featuring character $\alpha$. We used the \texttt{scikit-learn} Python machine learning package to perform NNMF.

\section{Results}

\subsubsection{Network Topology and Narrative Structure}
\begin{figure*}
\includegraphics[width=15cm]{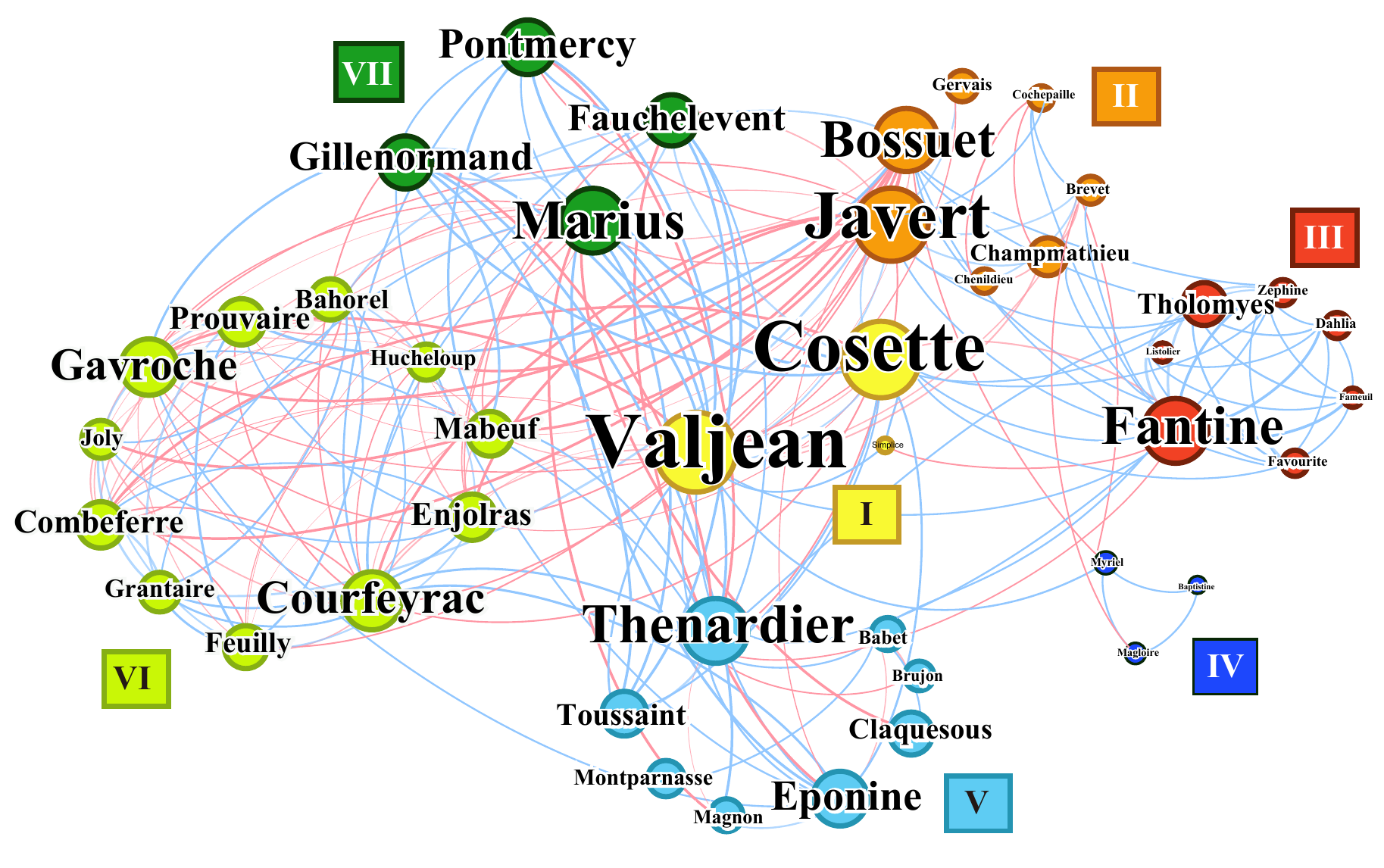} 
\caption{{\bf The Character Network of~\lesmis~and Its Community Structure.}
The character network of~\lesmis.  The node radius is proportional to its degree (the number of its neighbors).  The network shows many common characteristics of a social network such as the small-world property and the community structure. The node color indicates the community to which it belongs (we identify seven, labeled I to VII), while the edge color indicates the sign of the cosentiments of the character pair (blue for positive, and red for negative), defined and discussed further in Sec.~\ref{sec:sentimentanalysis}.}
\label{fig3}
\end{figure*}

This section is a summary of our previous work~\cite{Min:2016aa}. We start by constructing the network of characters based on Fig.~\ref{fig1}. The final network of \lesmis~contains 63 characters after very minor ones are excluded.  Drawing an edge between two characters if they have appeared in a chapter together results in $m=504$ edges. The network is shown in Fig.~\ref{fig3}. In it, $25.8\%$ of the character pairs are connected, the mean geodesic length is $1.85$, the network diameter is $4$ (between the pair of Babet and Geborand, and 17 other pairs of relatively minor characters), and the clustering coefficient is $0.77$~\footnote{Although our network appears denser than typical social networks~\cite{wasserman1994social,marsden1990network}, this is likely due to the fact that most characters of the novel are involved in some common plot while the rest of the story world is pushed into the background.}.

\begin{figure*}
	\includegraphics[width=12cm]{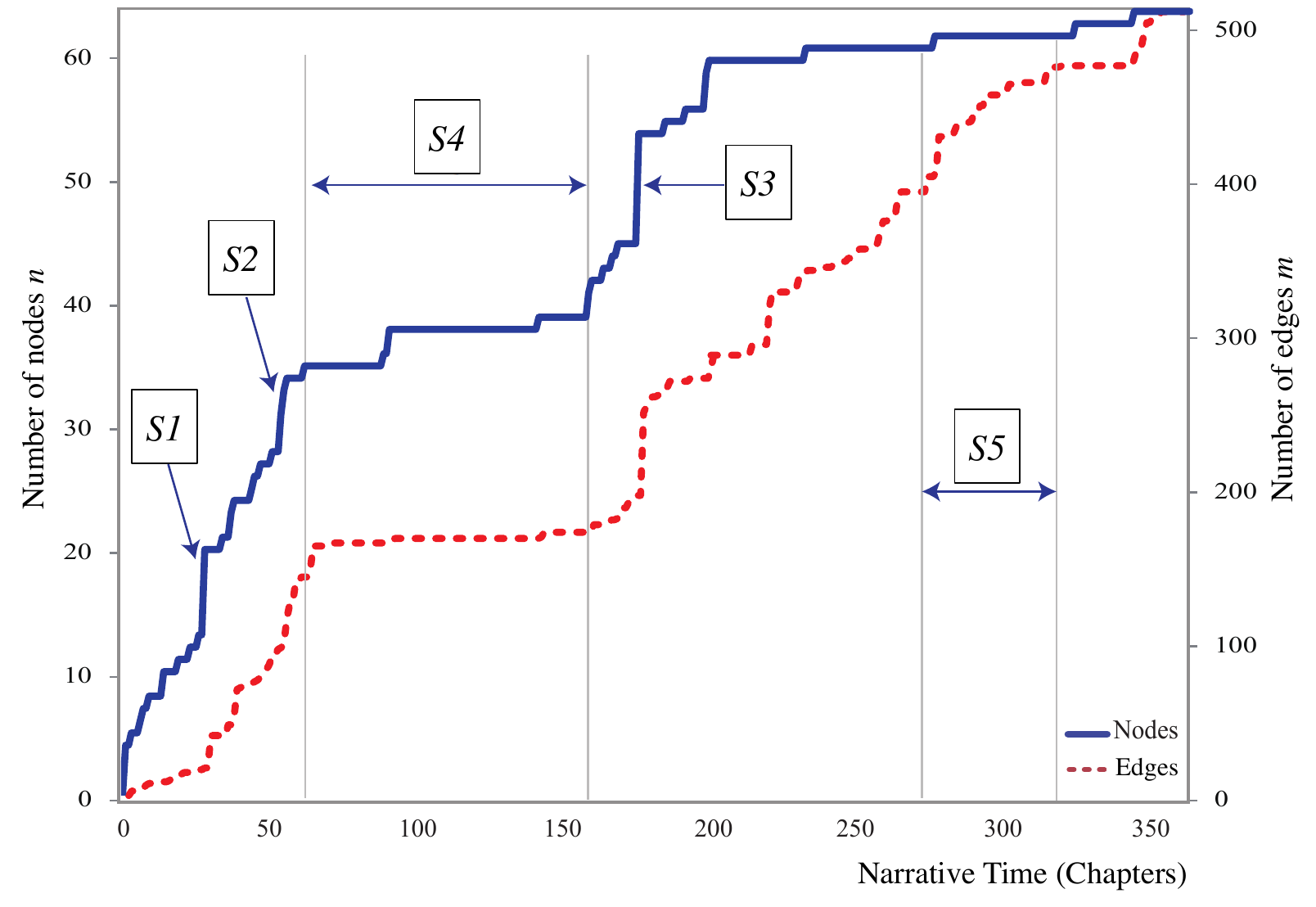}
	\caption{{\bf Growth Patterns of the Character Networks} The number of characters $n$ and the number of edges $m$ in~\lesmis~grow in a nonlinear fashion, indicating that different stages in narratives contribute differently to the network growth via character introduction or formation of new connections.}
	\label{fig4}
\end{figure*}

Based on the distinction between different stages in a narrative, we can assume that $n$ and $m$ would not simply increase linearly in time but nonlinearly in accordance with the nature of the stages. In Fig.~\ref{fig4} we show the growth of $n$ and $m$ along the narrative time measured in chapters.  As expected, the growth is not linear, especially for the number of nodes $n$. After the first batch of characters are introduced at the beginning of the narrative, there are specific points in the narrative where many new characters are introduced simultaneously (noted \emph{S1}, \emph{S2}, and \emph{S3} in Fig.~\ref{growth}) that suggest they are the Exposition stages.  An inspection of the actual story confirms this:
\begin{itemize}
	\item Stage \emph{S1}: Fantine's friends are introduced as her happy days are depicted.
	\item Stage \emph{S2}: Valjean's former fellow prison inmates testify during the trial of the fake Valjean.
	\item Stage \emph{S3}: ``The Friends of ABC'' (young progressives) are introduced, shown debating various social issues of the day.
\end{itemize}

There is also a stretch of chapters (\emph{S4}) where the network shows little growth. This part largely coincides with Volume 2 (``Cosette'') of the novel composed of those chapters that contain no narrative progression (i.e. the author digresses to discuss the battle of Waterloo, religion, the vagrant children of Paris,~\etc) or that show no network growth, being mainly about Valjean and Cosette's flight from the pursuit of Thenadier while avoiding people in general. Finally, near the end of the narrative at $\emph{S5}$, it is the number of edges $m$ that lead the growth of the network while $n$ shows little increase. This -- new edges being created between existing nodes without the addition to new ones -- implies a convergence of the characters into a common environment: this part in fact describes the scene at the barricade where nearly all major characters (who have been introduced before) converge.

\begin{figure*}
	\includegraphics[width=16cm]{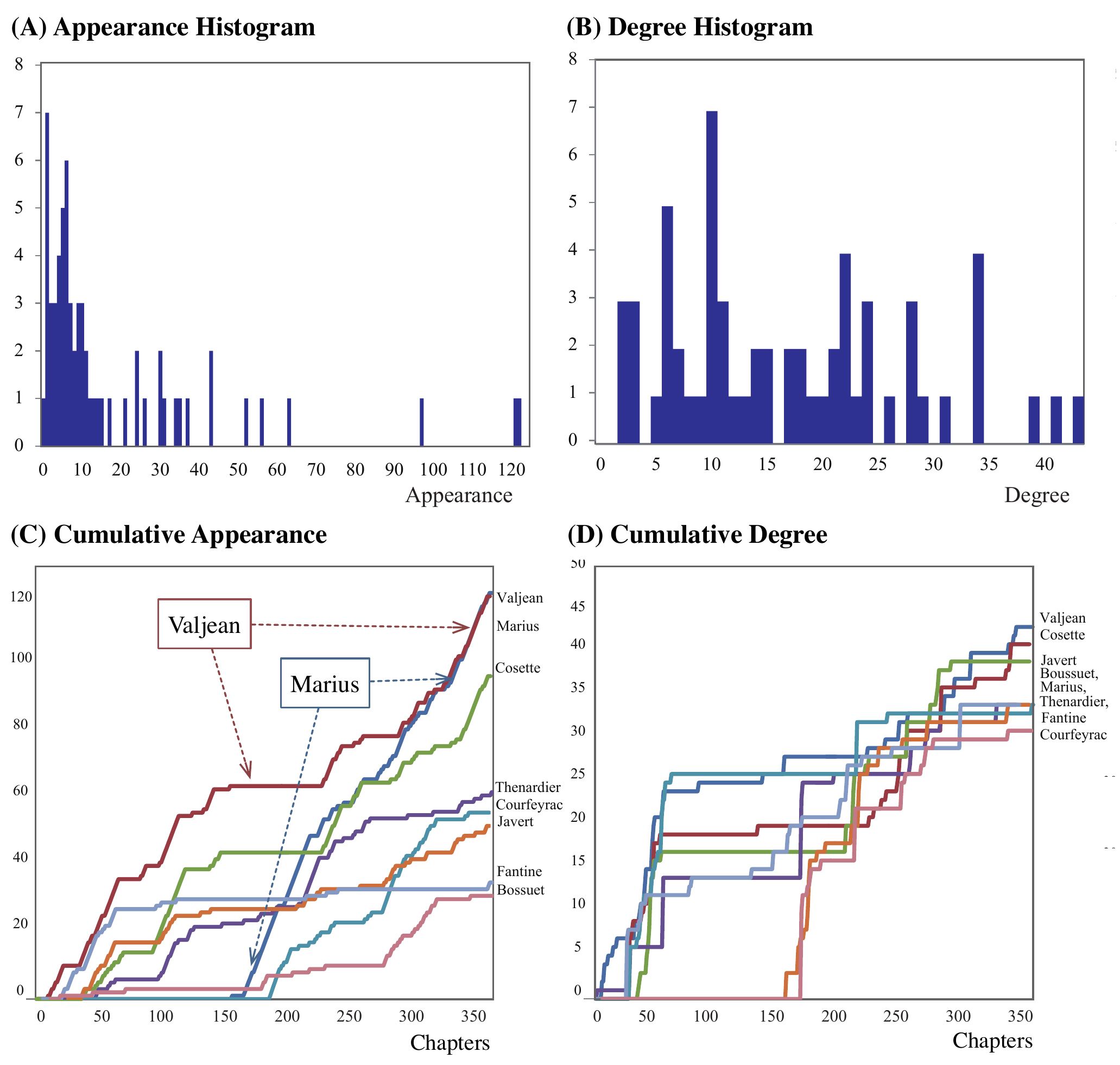} 
	\caption{{\bf Centralities of Characters in~\lesmis} Histograms of (A) appearance and (B) unweighted degrees of the characters of~\lesmis. The histograms are relatively skewed, with some characters having high values and many having small values. The three most frequently appearing characters are Marius (122), Valjean (121), and Cosette (97), while the three highest-degree characters are Valjean (43), Cosette (41), and Javert (39), and the three highest-weighted degree characters are Valjean (5203), Marius (4148), and Cosette (3977). The discrepancies indicate the differences in the characteristics of their social networks. (C) and (D) are the growths of these quantities for each character, showing the differing points at which the characters are actively depicted.} 
	\label{figcents} 
\end{figure*}

In Fig.~\ref{figcents} we show the appearance $a$ and degree $k$ of the individual characters. The final histogram of $a$ is shown in Fig.~\ref{figcents}~(A).  It has a skewed distribution with many characters appearing in a handful of chapters and a few characters appearing in many chapters, for instance Marius appears in $122$ chapters, Valjean in $121$, and Cosette in $97$, whereas the mean and the median are $19.3$ and $9$ respectively, nearly an order of magnitude smaller than the most frequent characters.  In Fig.~\ref{figcents}~(C) we show the temporal growth of each character's cumulative appearance. Although Marius and Valjean are similar in the total appearances ($122$ and $121$, respectively), how these values are reached are very different.  Valjean first appears in the beginning of the novel, then with regularity until there is a noticeable absence between chapters $160$ and $233$ (indicated by a plateau).  During Valjean's absence, Marius, making his first appearance in chapter $170$, takes the center stage in the novel and appears in almost every chapter until he overtakes Valjean in appearance.  This is a direct reflection of the structure of~\lesmis: the first part is mainly about Valjean (with Marius absent), the second part is mainly about Marius (with Valjean absent), and the final part features both as major characters.  The degree $k$ (Fig.~\ref{figcents}~(B)~and~(D)), on the other hand, differs in interesting ways from $a$. The three highest-degree nodes are Valjean ($k=43$), Cosette ($k=41$), and Javert ($k=39$), whereas Marius is down to $k=34$.  The degree therefore captures the nature of the social sphere around a character that appearance alone cannot tell: Valjean is a well-travelled character linking many different spheres of the story world, whereas Marius associates with a narrow pool of characters (namely the young fellow rebels) and his love interest Cosette.

\subsubsection{Sentiment Analysis and Narrative Progression}
\label{sec:sentimentanalysis}
The chapter sentiments are shown in Fig.~\ref{fig2}~(C).  We also study how the sentiments and content of chapters match: Positive chapters tend to depict uplifting characters (e.g. Myriel, a virtuous man) and events (e.g., Fantine going on a picnic, Cosette and Marius falling in love,~\etc), whereas negative chapters depict pain and suffering (e.g., Valjean nearly drowning, Fantine in misery, war, lovers parting,~\etc).   We also see alternating clusters of positive and negative chapters, indicating a certain pattern of emotional fluctuations.  This is reminiscent of an interpretation of narrative as a metaphor for life that fluctuates between contradictory states of harmony and peace, and tension and fear~\cite{mckee1997substance}.   We also note that the average chapter SPI is $\overline{\sigma}=0.06\pm 0.01$, i.e. net positive.  We believe this is an example of the so-called ``Pollyanna effect'' referring to a universal positivity bias in human language~\cite{dodds2015human}.  

\begin{figure*}
\includegraphics[width=15cm]{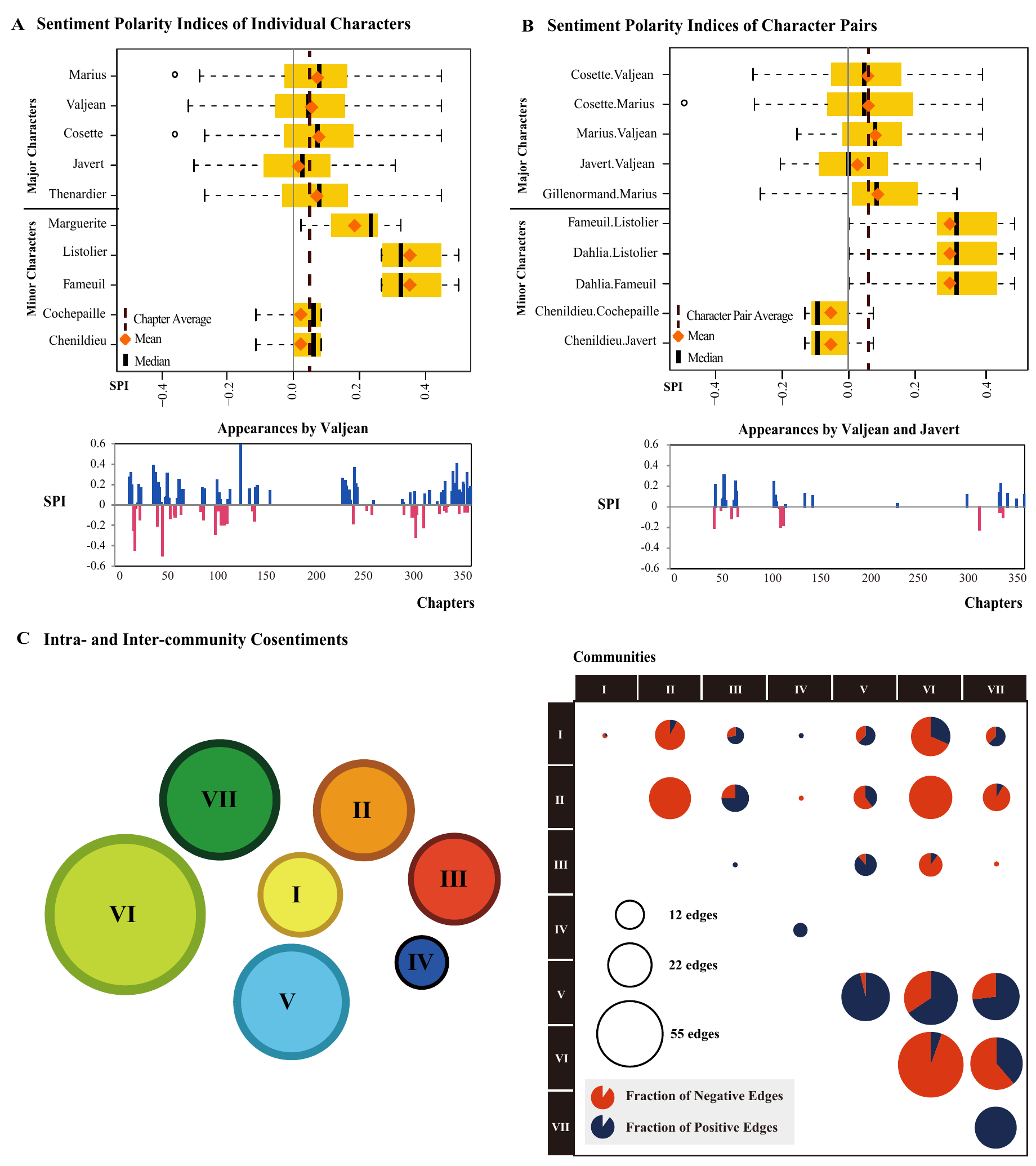}
\caption{{\bf Sentiments of Characters, Character Pairs, and Communities.}
(A) Sentiment Polarity Indices (SPIs) for the characters of~\lesmis. The yellow boxes indicate the SPI ranges of 50\% of the chapters around each character's median (25\% below, 25\% above).  The leading characters (higher in the plot) feature a wider range of SPIs than the marginal ones (lower in the plot), reflecting their role in the sentimental fluctuations of the narrative.  The SPIs of the chapters in which Valjean appears are shown below. (B) SPIs for character pairs. Valjean indeed shows an higher SPI when together with prot\'eg\'ee Cosette than pursuer Javert, although SPIs for leading characters again show a wide range. (C) The intra- and inter-community cosentiments. Communities II and VI are in general the most negative inside, due to the fact that  prisoners and revolutionaries share difficult and tragic experiences (harsh prison terms and death at the barricade). Communities V and VII are the most positive inside.  Between communities, II and VI are the most negative, due to Javert's presence at the barricade with the revolutionaries.}
\label{fig5}
\end{figure*}

We show the sentiments $\sen$ for select characters and pairs in Fig.~\ref{fig5}.  In Fig.~\ref{fig5}~(A) we show ten characters -- five major (frequently appearing) and five minor (infrequently appearing) -- for comparison.  While their average values are positive (due to the Pollyanna effect), the joyless Javert is more negative than other main characters such as Marius, Valjean, and Cosette. Nevertheless, major characters experience a wider range of SPIs than the minor ones, which we believe indicates their sentimental complexity. In the figure we see that Valjean appears frequently in both positive and negative chapters, showing his role as the carrier of varying sentimental states, in contrast to short-lived minor ones.  In Fig.~\ref{fig5}~(B) we show the SPIs of a number of character pairs.  Valjean understandably shows a higher average SPI when with his adoptive daughter Cosette than with his archnemesis Javert, although the wide range of SPIs again indicate the sentimental complexity of the leading character pairs. The Pollyanna effect still stands true here; in general, the average SPI of character pairs (dotted line) is a positive value at $\overline{\sen_0}=0.07$.  Therefore it is sensible to define the \defn{cosentiment} of a character pair $(\alpha,\beta)$ to be $\overline{\sen}[\alpha,\beta]-\overline{\sen_0}$. This quantity was already used for edge colors in Fig.~\ref{fig3}.  We can also use this to study the sentimental states within and between communities, shown in Fig.~\ref{fig5}~(C). In the figure, the diagonal elements show the fractions of positive and negative edges inside the communities, whereas the off-diagonal elements show those between two communities. The circle radius indicates the logarithm of the number of edges.  Communities II and VI are in general the most negative inside, showing the harsh and tragic nature of the common experiences of the prisoners and revolutionaries. To the contrary, Communities V and VII are the most positive inside.  Between communities, II and VI are the most negative, due to Javert's presence at the tragic barricade scene with the revolutionaries.

\begin{figure*}
\includegraphics[width=16cm]{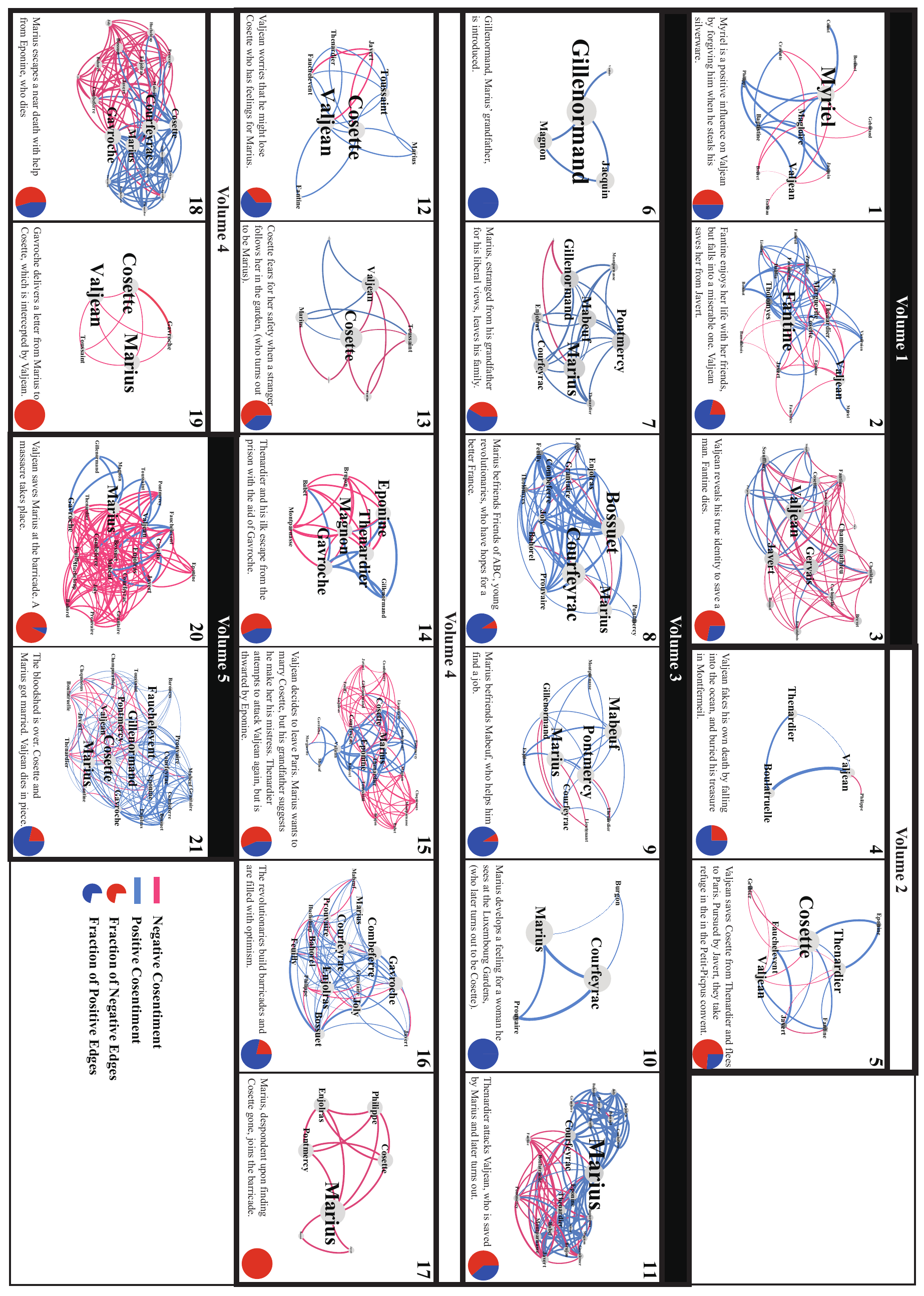}
\caption{{\bf Network Snapshots Showing Sentiments and Narrative Flow}
Snapshots of character networks in the 21 Sequences of \lesmis.  Edges are colored according to the cosentiment between the characters. The fractions of positive and negative edges are indicated in each snapshot, along with the summary of major plots in the Sequence.  The sentimental fluctuations often reflect the build up of drama, tension, and resolution.}
\label{fig6}
\end{figure*}

We now study the sentimental qualities of the network and how they change along the narrative progression. It is shown in Fig.~\ref{fig6}, where each panel corresponds to a Sequence of the novel first introduced in Fig.~\ref{fig1}~(B).  The definition and rationale for the Sequence are as follows:  Sometimes a plot or a storyline may span multiple consecutive narrative units, which makes it reasonable to bundle them into a larger one. To achieve it we need to determine the similarity between subsequent narrative units.  One possibility we use here is the character composition; consecutive units belonging to the same or highly similar plots are likely to contain similar characters.  Specifically, starting from the 40 Books of~\lesmis (excluding eight that contain no characters), we bundle the consecutive ones whose characters are similar above a prescribed threshold.  Using the cosine similarity (although others such as the Jaccard index may be used) and setting the threshold to be the average similarity ($0.49$) between consecutive book pairs, we end up with the $21$ Sequences shown in Fig.~\ref{fig6}. We also show the fraction of negative and positive edges.

The correlation between sentimental fluctuations and narrative flow are perhaps the best understood from Fig.~\ref{fig6} by studying Marius and his revolutionary friends.  When they are first introduced in Sequence 8, the sentiment is overwhelmingly positive, reflecting the air of optimism from their cause.  Such initial positivity is not long-lived, however, as they have to struggle with their adversaries in subsequent Sequences 11, 14, and 15.  After they overcome these challenges they briefly regain their positive sentiment (Sequence 16), but then are thrust into the most tragic and climactic circumstances (Sequences 17--20) that show high negativity. Finally, at the end of the novel (Sequence 21) the resolution is reached showing a highly positive sentiment. The fluctuations between positive and negative in this fashion are known to be by design~\cite{mckee1997substance}

\subsubsection{Topic Modeling and Mapping Interaction Dynamics via Topical States}
\begin{figure*}
\includegraphics[width=15cm]{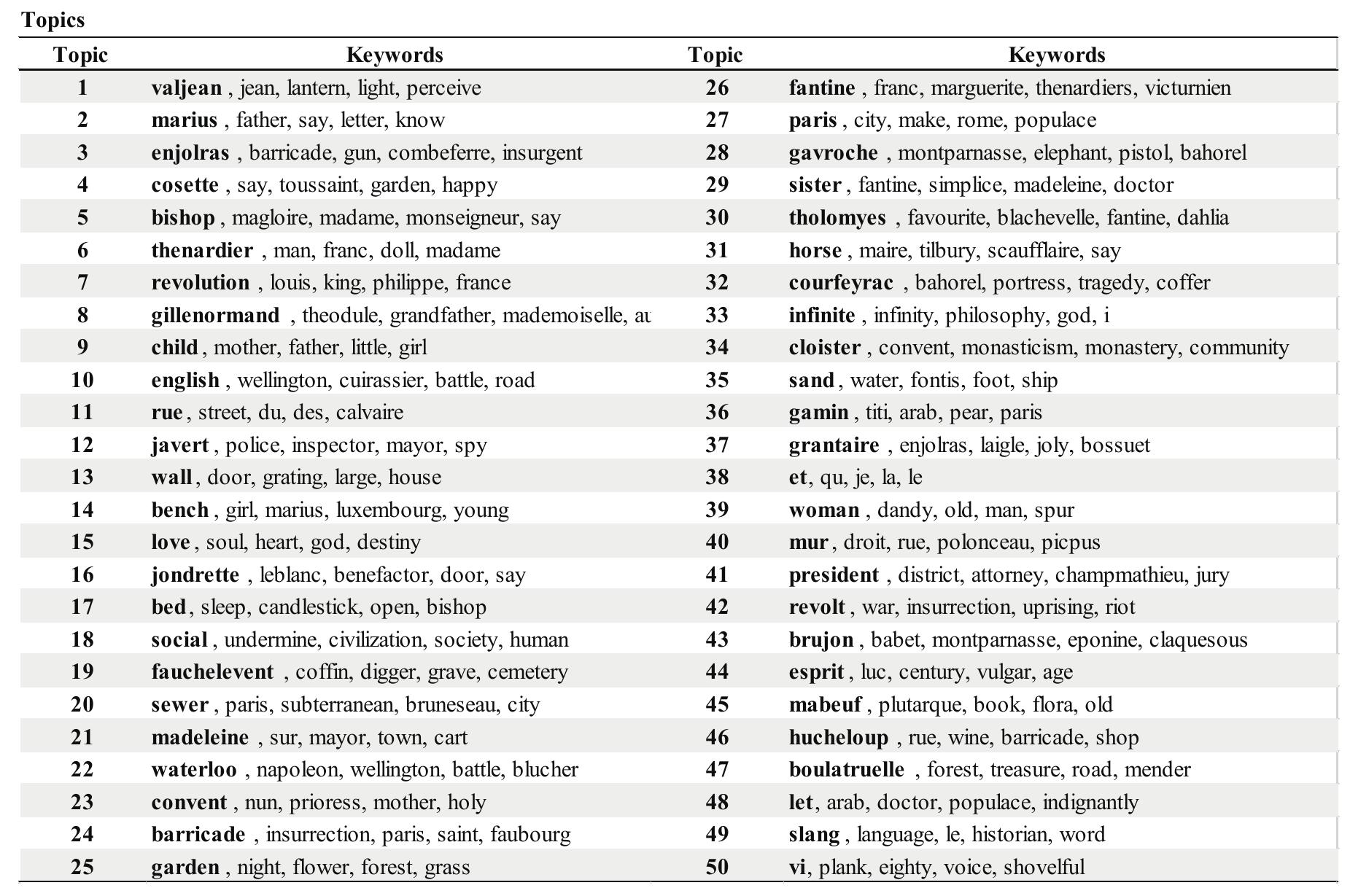} 
\caption{{\bf Complete List of Topics for \lesmis.}
50 Topics of \lesmis~found via Non-Negative Matrix Factorization (NNMF). Strongly associated keywords are also listed (the strongest keywords in bold). The topics are frequently about the characters (e.g. T1, T2, and T3), places (e.g. T11, T20, and T25), or events (e.g. T7, T22, and T42).}
\label{fig7}
\end{figure*}
\begin{figure*}
\includegraphics[width=15cm]{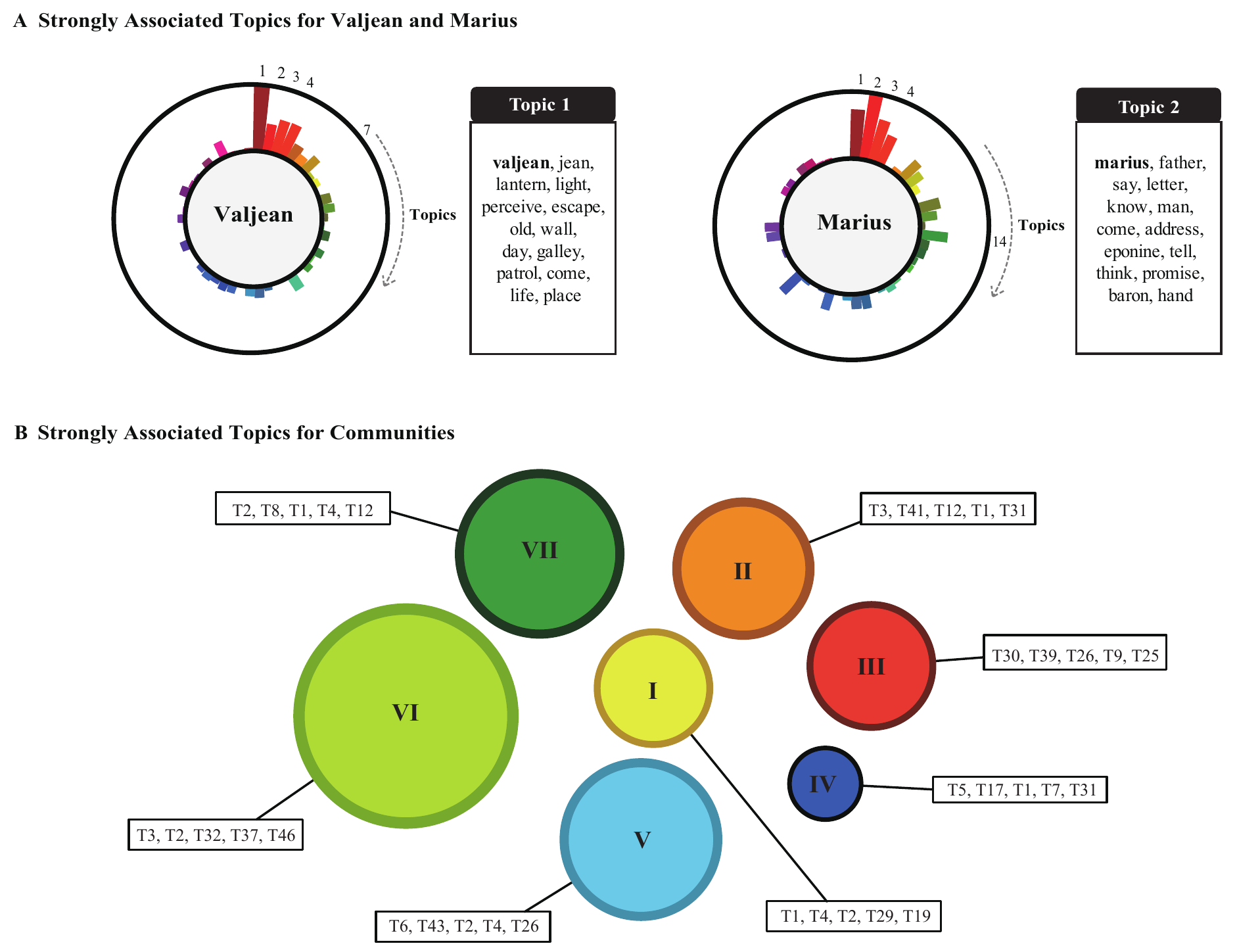}
\caption{{\bf Topics Associated with Valjean, Marius, and The Communities.}
(A) Topics strongly associated with Valjean (left) and Marius (right). The topic-character association strengths are scaled so that the largest value fills the space between the two circles. Topic T1 is the most relevant to Valjean, while Topic T2 is to Marius. They contain the respective character names as the strongest keywords, but also contain with words closely related to each character. (B) Topics strongly associated with the communities in Fig.~\ref{fig2}. The topics can be about the characters inside the community, or even from the outside as long as they are sufficiently associated with multiple members of the community. For instance, T2 (\kw{marius}), T29 (\kw{sister}, \kw{fantine}), and T19 (\kw{fauchelevent}) are strongly associated with Community I, although the characters belong to other communities. The topics can also be the events involving the community members, for instance T41 (the trial -- \kw{attorney}, \kw{jury}) for Community II composed of Javert and Valjean's fellow prison inmates.}
\label{fig8}
\end{figure*}

We set $\dimT=50$. The results for all 50 are given in Fig.~\ref{fig7}.  The keywords (the strongest ones are in bold) tell us that the topics are often about the characters (e.g. T1, T2, and T3), places (e.g. T11, T20, and T25), or events (e.g. T7, T22, and T42).  The character--topic associations $\set{t_{\alpha k}}$ are visualized in Fig.~\ref{fig8}~(A) for Valjean and Marius, scaled so that the strongest topic fills the space between the two circles. The five strongest topics for each characters are T1, T4, T3, T2, and T7 for Vajean, and T2, T1, T3, T4, and T14 or Marius.  From Fig.~\ref{fig7} we see that they are about themselves and related characters or actions (\kw{valjean}, \kw{escape}, \kw{marius}, \kw{eponine}, \etc).  We can also use them to identify topics associated with the communities by summing up the $t_{\alpha k}$ over the chapters that contain two or more of the members of the community, which are shown in Fig.~\ref{fig8}~(B).  The topics shown are relevant to multiple members of the group, for instance, characters from inside the community (e.g., T1 and T4 for Community I) or  outside (e.g., T2 and T29  for Community I), or the events or places, for instance T41 (the trial) for Community II of Javert and Valjean's fellow prison inmates.

We now introduce an interesting use of topics for representing narrative dynamics. An impactful event in a person's life is one that brings about significant changes in the person's state.  This means that even in a narrative, if one could define a character's state at a given point, one could measure the impact or significance of an event by comparing the states from before and after the event.  We use topic modeling to do exactly this, by interpreting $t_{\alpha k}$ as the \defn{topical state} of the character. The idea is straightforward: Since an associated topic indicates the action, events, interactions,~\etc~taking place in the character's presence, it can be understood as telling us the situation or the state of the character.  While Fig.~\ref{fig8}~(A) shows the topical states averaged over the entire novel, we can define a character's topical state at a given point in the narrative by obtaining the topical associations from the corresponding chapter(s).   As an example now study the  impact that the interactions between Marius and Valjean have on the character's states. For simplicity we consider Valjean and Marius to be interacting largely two times in \lesmis, prompting us to partition the novel into the following four phases:
\begin{enumerate}
	\item Phase I (Chapters 1 to 233): Before the first interaction. Valjean and Marius lead separate lives.
	\item Phase II (Chapters 234 to 266): The first interaction take place.  Marius falls in love with Cosette, causing Valjean to become anxious about losing her.
	\item Phase III (Chapters 272 to 295): Valjean is absent from the narrative, so no interaction takes place. Marius parts from Cosette, then joins the revolutionaries at the barricade.
	\item Phase IV (Chapters 296 to the end of the narrative): The second interaction takes place. Marius gets injured at the barricade, then is rescued by Valjean. Cosette and Marius marry. Valjean dies.
\end{enumerate}

\begin{figure*}
\includegraphics[width=18cm]{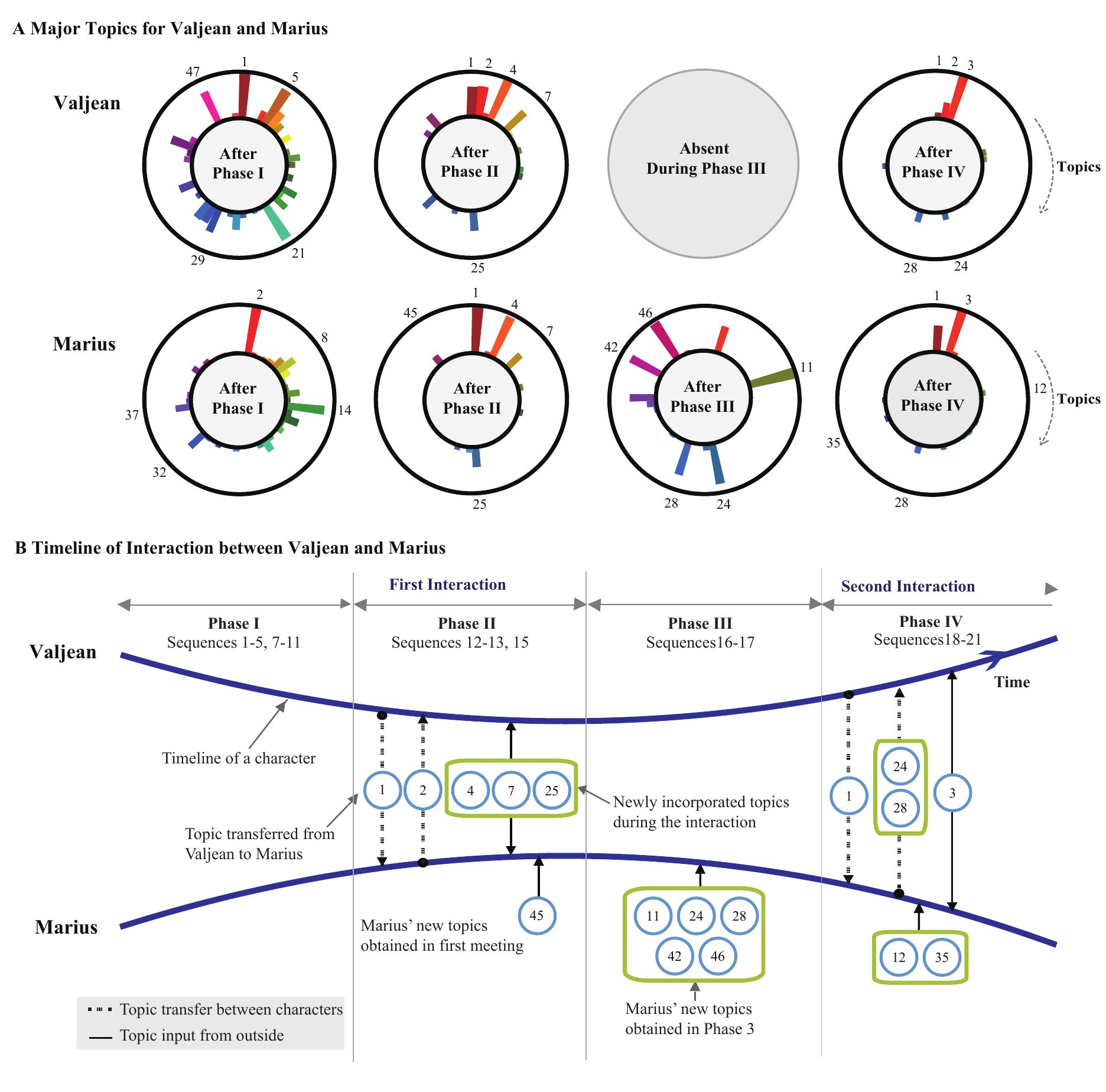} 
\caption{{\bf Mapping Out Interactions Diagrammatically As Dynamic Topic Exchange.}
Events lead to character transformation, which we quantify via the character's topical states.  With respect to the interaction between Valjean and Marius, we divide \lesmis~into four phases. (A) The net changes in the topical states of the characters at the end of each phase, quantified by the differences $t_{\alpha k}$. After Phase II, T2 (the strongest topic for Marius before) shows a sharp increase for Valjean. Likewise, T1 (Valjean's strongest topic before) shows a sharp increase for Marius. T4, T7, and T25 increase for both characters, while T45 increases only for Marius. (B) Diagrammatic representation of the changes of Marius' and Valjean's topical states as 'topic transfers' during each phase; Topics can be exchanged between characters (e.g., T1 and T2 during Phase II) or enter either character's topical state exogenously (e.g., T4, T7, T25, and T45). The dirctions can also reflect those of actual story elements: during Phase IV (Chapters 296--365), Valjean, prompted by a letter from Marius, joins the barricade. This is directly reflected in the transfer of topics T24 and T28 from Marius to Valjean.}
\label{fig9}
\end{figure*}

Our strategy now is to observe the changes in characters' states $t_{\alpha k}$. We then use them to understand the details of the interaction dynamic. First, the changes in $t_{\alpha k}$ for the characters at the end of each phase are shown in Fig.~\ref{fig9}~(A), obtained by subtracting the $t_{\alpha k}$ immediately before the interactions from that from immediately after. At the end of Phase I, Valjean is the most strongly associated with T1, T5, T21, T47, and T29, whereas Marius is with T2, T14, T8, T32, and T37 which represent their trajectories up to that point according to Fig.~\ref{fig7}.  They share no common topics, as expected from the lack of any interaction up to that point -- in fact, the correlation between their $\set{t_{\alpha k}}$ is negative at $-0.20\pm 0.01$.  At the end of Phase II after their first interaction the correlation increases to $0.42\pm 0.01$, showing that an interaction works to correlate the character states. At the end of Phase III (no interaction) it decreases again slightly to $0.33\pm0.001$.  At the end of Phase IV where they interact again for the final time and quite extensively it reaches its highest value of $0.70\pm0.01$. These show that an interaction functions to assimilate the characters' states, and an inspection of the changes $\Delta t_{\alpha k}$ provides us with more detail of this assimilation dynamics.  For simplicity, we again focus on the five topics (for each character) that gain the most in strength after each phase, shown in Fig.~\ref{fig9}~(A). After the first interaction, we find that the five such topics for Valjean are T4, T2, T1, T7, and T25, whereas for Marius they are T1, T4, T25, T7, and T45. When we compare the strongly associated topics from before and after the interactions, we find there are some that we can interpret as having been \emph{transferred} from one character to the other. An example is T2 (\kw{marius}), the strongest one with Marius before Phase II, which gains the most for Valjean after. The same goes for T1 (\kw{valjean}), this time from Valjean to Marius.  Second, there are topics that have entered the characters states exogenously, i.e. those that not strongly associated with either character.  They represent new common experiences or interests that occur during the interactions: T4 (\kw{cosette}), T7 (\kw{revolution}), and T25 (\kw{garden}) are such cases. They again reflect the story accurately: Cosette becomes the focal point of both characters, as a new love interest for Marius that causes severe anxiety to Valjean.  Some topics enter only one character's state, such as T45 (\kw{mabeuf}) which is about a character Mabeuf who shares his story with Marius at the barricade, but has little to do with Valjean -- Valjean's topical state indeed has near-zero component of T45.  Next, during Phase III, T11 (\kw{rue}), T24 (\kw{barricade}), T46 (\kw{hucheloup}), T42 (\kw{revolt}), and T28 (\kw{gavroche}) gain the most strength for Marius, reflecting the events and the characters he experiences during that time.  Valjean is absent.  Finally, during Phase IV, T3 (\kw{enjolras}), T2, T28 (\kw{gavroche}), T24 (\kw{barricade}), and T1 gain the most strength with Valjean, whereas topics T3 T1, T28, T35 (\kw{sand}), and T12 (\kw{javert}) gain the most strength with Marius.  Note how the directionality of T28 and T24 from Marius to Valjean reflects the actual way things happen between the characters: Gavroche (T28), a friend of Marius', carries a letter from Marius to Valjean that motivates Valjean to join the barricade (T24) in search of Marius.  Our discussion here about topic transfers and entry can be systematically visualized as in Fig.~\ref{fig9}~(B) on top of the basic interaction timeline first introduced in Fig.~\ref{fig1}, showing that the textual information indeed allows to construct a much more detailed picture of an interaction than a simple occurrence-based network construction.

The results provided in this section, by showing that the story of a narrative can be identified, quantified, analyzed, and visualized by making use of appropriate analytical and computational tools, we believe demonstrate the benefits and opportunities of approaching traditional subjects as narrative from a novel perspective that allows us to find new patterns and gain a richer understanding not readily available previously.

\section{Discussions and Conclusions}
\label{sec:discussion}
In this paper we proposed a network-based framework for modeling a narrative by focusing on the characters and their interactions. We started by representing a narrative as a set of interacting character timelines, from which we constructed a growing character network. To legitimize our approach it was necessary to understand how the character network topology and dynamics reflected the narrative structure correctly.  We found that character centralities captured the role and the nature of the social spheres of characters in the narrative, while the temporal growth of network showed distinct phases with differing patterns of increasing nodes or edges depending on whether the narrative was focusing on isolated characters (stagnant growth), expanding the story world by introducing new characters (growth led by number of nodes), or when existing characters converge into the building process to the resolution (growth led by number of edges).

An important characteristic of well-written drama is that it evokes emotion in the reader, which in the western literary tradition is conventionally represented by the generic division of drama into comedy and tragedy. This had an interesting connection to a modern computational methodology called sentiment analysis. We found that many characters, especially the central ones, showed significant fluctuations of sentiments during the narrative flow, acting as the carriers of mood and emotions of the narrative.  This was true of character relationships as well, and we showed how the sentimental fluctuations correlated with the narrative progression that showed detectable patterns of dramatic tension build-up and resolution.

Finally, we used topic modeling as a way to define the state of a character via the topics (keywords) with which they are associated at various points in the narrative. This allowed us to trace quantitatively the changes in characters' states, and quantify and map out the details of an event or an interaction between characters. We also demonstrated that the flow of topics between characters can reflect the actual story in interesting ways, providing us with a way to systematically represent the patterns of character interactions that previously resided in the text of the narrative.

We believe that our paper presents a wide range of ideas for studying narrative structures that merit further exploration using the methods of network science, data analysis, and computational linguistics.  Looking further, representing a narrative as a dynamically unfolding system of character networks and interactions also sets the stage for using theories and tools for understanding of dynamical systems, not only networks. Advances in this area have practical implications as well, such as an improved algorithm for computer-assisted writing and storytelling which no doubt can benefit from a more robust understanding of the patterns of character relationships and interactions.  Given the ubiquity and importance of narratives, we hope that future developments based on our work will be beneficial for a wide range of fields including literature, communication, and storytelling.

\acknowledgements
The authors would like to thank Kyungyeon Moon, Wonjae Lee, and Bong Gwan Jun for helpful comments. This work was supported by the National Research Foundation of Korea (NRF-20100004910 and NRF-2013S1A3A2055285), BK21 Plus Postgraduate Organization for Content Science, and the Digital Contents Research and Development program of MSIP (R0184-15-1037, Development of Data Mining Core Technologies for Real-time Intelligent Information Recommendation in Smart Spaces).


\end{document}